\documentclass[10pt,twocolumn,letterpaper]{article}

\usepackage{cvpr}
\usepackage{times}
\usepackage{epsfig}
\usepackage{graphicx}
\usepackage{amsmath}
\usepackage{amssymb}

\usepackage{color}

\usepackage{subfigure}

\usepackage{interval}

\usepackage{caption}

\usepackage{pifont}

\usepackage{makecell}
\usepackage{booktabs}
\usepackage{multirow}
\usepackage{siunitx}

\usepackage[T1]{fontenc}
\usepackage{fix-cm}

\newcolumntype{?}{!{\vrule width 1pt}}
\newcolumntype{C}[1]{>{\centering}m{#1}}

\newcolumntype{X}{@{\hskip\tabcolsep\vrule width 1.5pt\hskip\tabcolsep}}

\usepackage{lipsum} 
\definecolor{orange}{RGB}{255,127,0}

\usepackage[pagebackref=true,breaklinks=true,letterpaper=true,colorlinks,bookmarks=false]{hyperref}

\usepackage{scrextend}
\usepackage{hyperref}       
\usepackage{url} 
\usepackage[dvipsnames]{xcolor}
\definecolor{mypink}{rgb}{0.858, 0.188, 0.478}
\definecolor{brightpink}{rgb}{1.0, 0.0, 0.5}
\definecolor{amethyst}{rgb}{0.6, 0.4, 0.8}
\hypersetup{
    colorlinks,
    urlcolor={brightpink!100!black}
}

\cvprfinalcopy 

\def\cvprPaperID{7757} 

\ifcvprfinal\fi
\begin{document}

\title{Classifying, Segmenting, and Tracking\\Object Instances in Video with Mask Propagation}

\author{Gedas Bertasius, Lorenzo Torresani\\
Facebook AI}

\maketitle
\thispagestyle{empty}

\begin{abstract}
We introduce a method for simultaneously classifying, segmenting and tracking object instances in a video sequence. Our method, named MaskProp, adapts the popular Mask R-CNN to video by adding a mask propagation branch that propagates frame-level object instance masks from each video frame to all the other frames in a video clip. This allows our system to predict clip-level instance tracks with respect to the object instances segmented in the middle frame of the clip. Clip-level instance tracks generated densely for each frame in the sequence are finally aggregated to produce video-level object instance segmentation and classification. Our experiments demonstrate that our clip-level instance segmentation makes our approach robust to motion blur and object occlusions in video. MaskProp achieves the best reported accuracy on the YouTube-VIS dataset, outperforming the ICCV 2019 video instance segmentation challenge winner despite being much simpler and using orders of magnitude less labeled data (1.3M vs 1B images and 860K vs 14M bounding boxes). 
\end{abstract}

\section{Introduction}

In this paper, we tackle the recently introduced video instance segmentation problem~\cite{Yang2019vis}. This task requires segmenting all instances of a predefined set of object classes in each frame, classifying them, and linking individual instances over the entire sequence. 


In recent years, convolutional networks have obtained remarkable results in still-image object detection~\cite{he2017maskrcnn,ren2015faster,girshick15fastrcnn,girshick2014rcnn}, and segmentation~\cite{long_shelhamer_fcn,DBLP:conf/cvpr/ZhaoSQWJ17,DBLP:journals/pami/ChenPKMY18,gberta_2016_CVPR}. However, extending these models to video instance segmentation is challenging. In order to localize objects precisely, these methods have to operate at very large spatial resolution. As a result, detectors based on the popular ResNet-101 or ResNet-152 backbones~\cite{He2016DeepRL} can rarely fit more than one image per GPU during training. In the context of video instance segmentation this is problematic because tracking objects over time requires analyzing multiple video frames simultaneously.

To address this issue, one could reduce the spatial resolution of the input and fit more video frames in a GPU. However, doing so typically leads to a significant drop in segmentation or detection performance. Alternatively, one could perform high-resolution instance segmentation on individual frames and then link segmentations temporally in a separate post-processing stage. However, performing instance segmentation and tracking in two disjoint steps often produces suboptimal results, because these two tasks are closely intertwined. The key challenge then becomes designing a unified model that can track objects in video while maintaining strong detection accuracy.

\begin{figure}[t]
\begin{center}
   \includegraphics[width=1\linewidth]{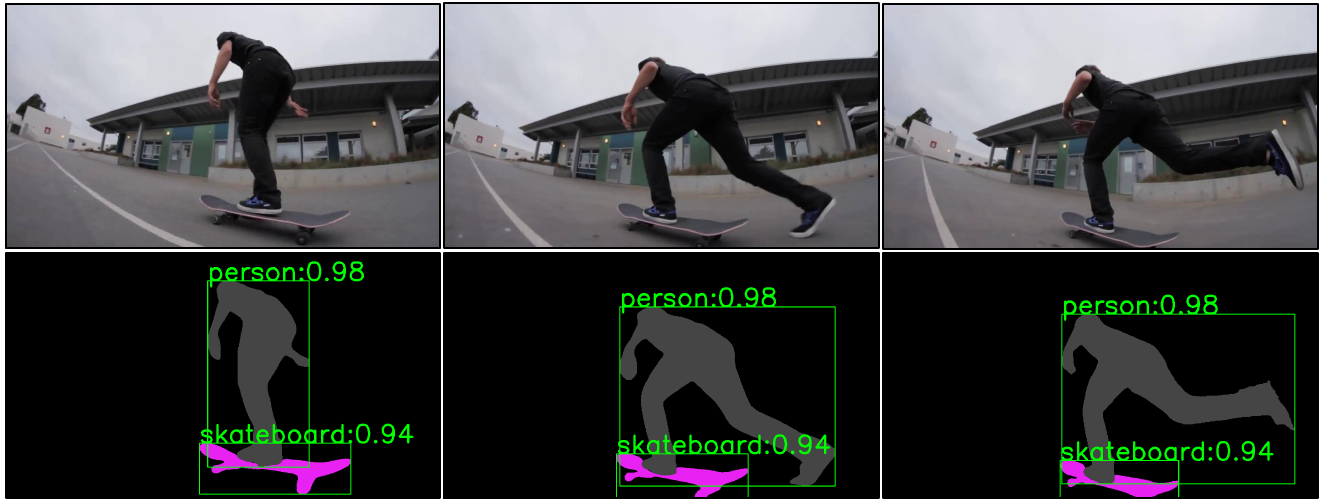}
\end{center}
\vspace{-0.4cm}
        \caption{In this paper, we tackle the problem of video instance segmentation, which requires classifying, segmenting, and tracking object instances in a given video sequence. Our proposed Mask Propagation framework (MaskProp) provides a simple and effective way for solving this task.\vspace{-0.6cm}}
\label{main_fig}
\end{figure}




\begin{table*}[t]
\vspace{-0.3cm}
\setlength{\tabcolsep}{3.0pt}
\footnotesize
\begin{center}
 \begin{tabular}{c c  c  c  c} 
 \hline
 & & MaskTrack R-CNN~\cite{Yang2019vis} & ICCV19 Challenge Winner~\cite{vis_challenge_winner19} & MaskProp\\
 \hline
  \multirow{5}{*}{Model} 
 & Classification & cls head & Mask R-CNN~\cite{he2017maskrcnn}, ResNeXt-101 32x48d~\cite{mahajan2018weaklysup} & cls head\\
 & Localization & bbox head & Mask R-CNN~\cite{he2017maskrcnn} & bbox head\\
 & Segmentation & mask head & DeepLabv3~\cite{DBLP:conf/eccv/ChenZPSA18}, Box2Seg~\cite{luiten2018premvos} & mask head\\
 & Tracking & tracking head & UnOVOST~\cite{ZulfikarLuitenUnOVOST}, ReID Net~\cite{HermansBeyer2017Arxiv,Osep19ICRA} & mask propagation head\\
 & Optical Flow & - & PWC-Net~\cite{Sun2018PWC-Net} & -\\
 \hline
  \multirow{4}{*}{Pre-training Datasets} & ImageNet~\cite{ILSVRCarxiv14} (1.3M images) &  \checkmark &  \checkmark &  \checkmark\\
  & COCO~\cite{502} (860K bboxes) &   \checkmark &  \checkmark &  \checkmark\\
  & Instagram~\cite{mahajan2018weaklysup} (1B images) & - &  \checkmark & -\\
  & OpenImages~\cite{kuznetsova2018open} (14M bboxes) & - &  \checkmark & -\\
  \hline
  \multirow{2}{*}{Performance} & video mAP & 30.3 & 44.8 & \bf 46.6\\
  & video AP@75 & 32.6 & 48.9 & \bf 51.2\\
  \hline
 \vspace{-0.6cm}
\end{tabular}
\end{center}
\caption{A table comparing our work to prior video instance segmentation methods~\cite{Yang2019vis,vis_challenge_winner19}. The ICCV 2019 Challenge Winner~\cite{vis_challenge_winner19} decomposes video instance segmentation into four different problems, solves each of them independently using ensembles of different models, and then combines these solutions. In contrast, our approach relies on a single unified model trained end-to-end. Despite being simpler, and using several orders of magnitude less pretraining data (1.3M vs 1B images and 860K vs 14M bounding boxes) our model achieves higher accuracy. Furthermore, compared to MaskTrack R-CNN~\cite{Yang2019vis}, our work yields a 16.3\% gain in mAP (46.6\% vs 30.3\%).\vspace{-0.4cm}}
\label{methods_table}
\end{table*}

Currently, the best method for video instance segmentation  is the ICCV 2019 challenge winner~\cite{vis_challenge_winner19}. It tackles video instance segmentation by dividing it into four problems: 1) detection,  2) classification, 3) segmentation, and 4) tracking. These four problems are solved independently using several off-the-shelf components and their respective solutions are combined and adapted to the video instance segmentation task. However, despite effective performance, such an approach is disadvantageous because it requires designing and tuning a separate model (or, in some cases, an ensemble of models) for each of the four tasks. This renders the approach costly and cumbersome. On the other end of the complexity spectrum, MaskTrack R-CNN~\cite{Yang2019vis} is a simple unified approach trained end-to-end but it achieves significantly lower performance ($30.3$ vs $44.8$ video mAP). 




To address the shortcomings of these prior methods we introduce MaskProp, a simple mask propagation framework for simultaneously classifying, segmenting and tracking object instances in video. Our method adapts the popular Mask R-CNN~\cite{he2017maskrcnn} to video by adding a branch that propagates frame-level instance masks from each video frame to other frames within a temporal neighborhood (which we refer to as a clip). This allows our method to compute clip-level instance tracks centered at each individual frame of the video. These densely estimated clip-level tracks are then aggregated to form accurate and coherent object instance sequences for the entire video, regardless of its length. This renders our approach capable of handling challenging cases of occlusions, disocclusions, and motion blur.  Our method achieves the best reported accuracy on the YouTube-VIS dataset~\cite{Yang2019vis}, outperforming the ICCV 2019 challenge winner~\cite{vis_challenge_winner19} despite being much simpler and using significantly less labeled data (1000x fewer images and 10x fewer bounding boxes). In Table~\ref{methods_table}, we compare our approach vs these prior methods in terms of accuracy and other characteristics.


\section{Related Work}

\noindent \textbf{Instance Segmentation in Images.} Compared to instance segmentation in images~\cite{dai2016instance,he2017maskrcnn,li2016fully,DBLP:conf/cvpr/ArnabT17,8099788,kirillov-2017,8237640}, the problem considered in this paper requires not only to segment object instances in individual frames, but also to determine instance correspondences across multiple frames. We leverage the Mask R-CNN model~\cite{he2017maskrcnn} for still-image instance segmentation and adapt it to track object instances in video.

\noindent \textbf{Object Detection in Video.} Object detection in video requires classifying and localizing objects in every frame of a given video. Most modern video object detection systems~\cite{zhu17fgfa,DBLP:conf/eccv/BertasiusTS18,xiao-eccv2018,feichtenhofer2017detect} implement some form of spatiotemporal feature alignment for improving object detection accuracy in individual video frames. However, these systems are typically not designed for tracking object instances. In contrast, our mask propagation produces clip-level instance segmentations rather than frame-level bounding boxes.



\begin{figure*}[t]
\begin{center}
   \includegraphics[width=0.72\linewidth]{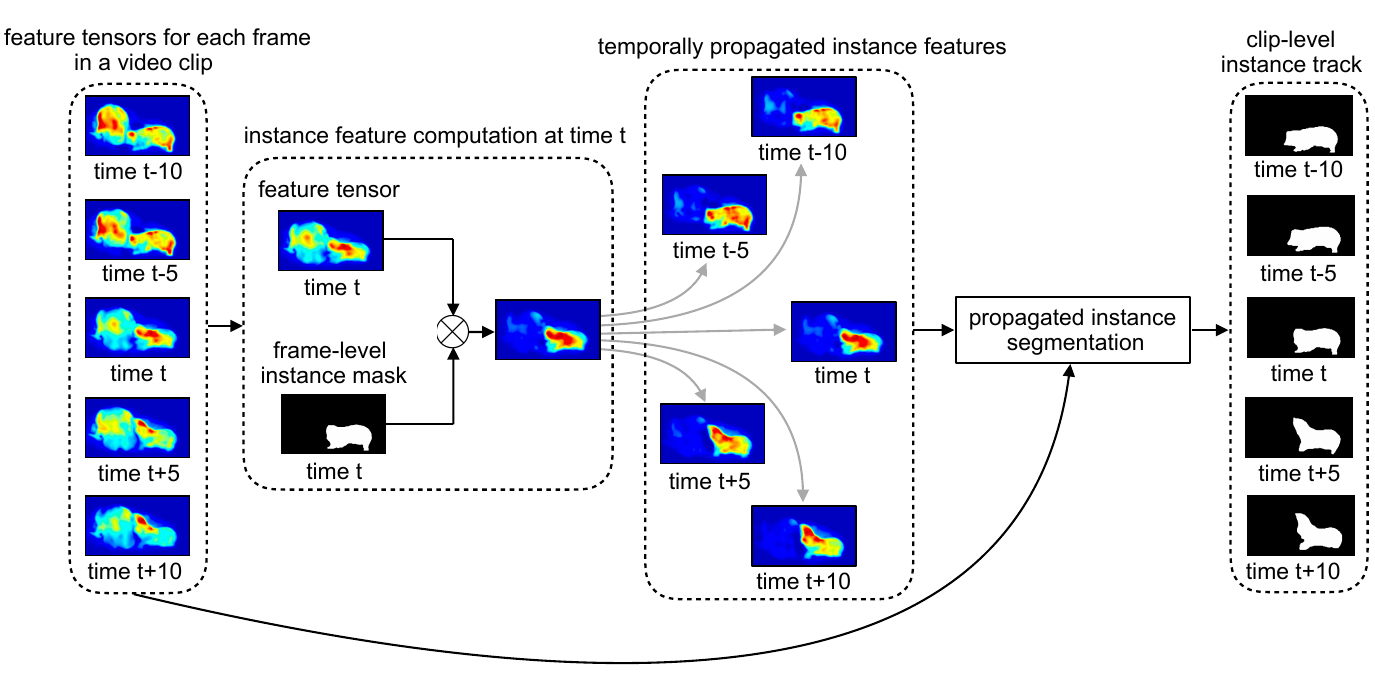}
\end{center}
\vspace{-0.4cm}
         \caption{An illustration of our MaskProp system, which takes as input a video clip centered around frame $t$, and outputs a clip-level instance track. Our mask propagation framework can be summarized in three high-level steps: 1)  An instance-specific feature map at time $t$ is computed by masking the frame features at time $t$ with the given instance segmentation for frame $t$ (one for each instance detected in frame $t$). 2) Next, we use our mask propagation mechanism to temporally propagate instance-specific features from frame $t$ to all the other frames in the clip. 3) Lastly, our model predicts instance-specific segmentations in every frame of the clip by implicitly matching the propagated instance features with the frame-level features computed at each time step. This last step yields clip-level instance tracks centered around frame $t$.\vspace{-0.3cm}}
\label{high_level_arch_fig}
\end{figure*}


\noindent \textbf{Video Object Segmentation.} The task of video object segmentation requires segmenting foreground objects in a class-agnostic fashion~\cite{DBLP:conf/cvpr/YangWXYK18,DBLP:conf/cvpr/JainXG17,8237742,Voigtlaender2019CVPR}, often by leveraging ground truth masks available for the first frame during inference~\cite{Cae+17,Perazzi2017,conf/cvpr/ChenPMG18,Herrera-Palacio:2019:VOL:3347450.3357662,feelvos2019}. Instead, video instance segmentation requires finding all instances of a predefined set of object classes in each frame, classifying them and linking them over the entire sequence. 

\noindent \textbf{Video Instance Segmentation.} The recently introduced video instance segmentation task~\cite{Yang2019vis} requires classifying, segmenting and tracking object instances in videos. This is the task considered in this work. There are only a few video instance segmentation methods we can compare our approach to. The MaskTrack R-CNN~\cite{Yang2019vis} presents a unified model for video instance segmentation. It augments the original Mask R-CNN~\cite{he2017maskrcnn} with a tracking branch that establishes associations among object instances segmented in separate frames. Furthermore, we include the ICCV 2019 video instance segmentation challenge winner~\cite{vis_challenge_winner19} in our comparison. This approach divides video instance segmentation into four separate subproblems: classification, detection, segmentation, and tracking. A separate model (or an ensemble of models) is used to solve each of these subproblems, and these solutions are then combined to produce video instance segmentation results. For brevity, from now on we refer to it as EnsembleVIS to indicate that it is an ensemble approach designed for video instance segmentation.

Our MaskProp framework provides advantages over both of these methods~\cite{Yang2019vis,vis_challenge_winner19}. Similarly to MaskTrack R-CNN~\cite{Yang2019vis}, our method is a unified and simple approach. However, our mask propagation branch is much more effective than the tracking branch of MaskTrack R-CNN, achieving much higher accuracy relative to this baseline. Furthermore, compared to  EnsembleVIS~\cite{vis_challenge_winner19}, our method 1) is much simpler, 2) uses significantly less labeled data, and 3) produces higher accuracy on YouTube-VIS~\cite{Yang2019vis}.

\section{Video Instance Segmentation}

\noindent \textbf{Problem Definition.} 
Let us denote with $V \in \mathcal{R}^{L \times 3 \times H \times W}$ an input video consisting of $L$ RGB frames of spatial size $H \times W$. The aim of our system is to segment and temporally link all object instances that are visible for at least one frame in $V$ and that belong to a predefined set of categories $\mathcal{C} = \{1, ..., K\}$ . To achieve this goal, our model outputs a video-level instance mask track $M^i \in \mathcal{R}^{L \times H \times W}$ with a category label $c^{i}  \in \{1, ..., K\}$ and a confidence score $s^{i} \in [0, 1]$ for each object instance $i$ detected in the video. 

\noindent \textbf{Evaluation Metric.} Video instance segmentation is evaluated according to the metrics of average precision (AP) and average recall (AR). Unlike in the image domain, these metrics are evaluated over the video sequence. Thus, to evaluate spatiotemporal consistency of the predicted mask sequences, the video Intersection over Union (IoU) between a predicted object instance $i$ and a ground truth object instance $j$ is computed as:

\vspace{-0.3cm}
\begin{equation}
\begin{split}
IoU(i,j) =\frac{\sum_{t=1}^T |M^i(t)  \cap \tilde{M}^j(t)|}{\sum_{t=1}^T |M^i(t)  \cup \tilde{M}^j(t)|}
\end{split} 
\end{equation}
where $\tilde{M}^j(t)$ is the ground-truth segmentation of object $j$ in frame $t$. To achieve a large IoU, a model must not only accurately classify and segment object instances at a frame-level, but also reliably track them over the video sequence. 

As in the COCO benchmark for image segmentation~\cite{502}, the metrics of AP and AR are computed separately for each object category, and then averaged over ten IoU thresholds from $50\%$ to $95\%$ at increments of $5\%$. Lastly, the resulting AP and AR metrics are averaged over the category set, which yields the final evaluation metric.

\section{Mask Propagation}
\label{sec:maskprop}


MaskProp takes a video $V$ of arbitrary length $L$ as input and outputs video-level instance segmentation tracks $M^i$, category labels $c^i$  and confidence scores $s^{i}$ for all objects $i$ detected in the video. In order to achieve this goal, our method first builds clip-level object instance tracks ${M}^i_{t-T:t+T} \in \mathcal{R}^{(2T+1) \times 1 \times H \times W}$ for each individual clip $V_{t-T:t+T} \in \mathcal{R}^{(2T+1) \times 3 \times H \times W}$ of length $(2T+1)$ in the video, i.e., for $t=1, 2, \hdots, L$ (clips at the beginning and the end of the video will include fewer frames).  

We want to use clips that are long enough to allow us to jointly solve instance segmentation and tracking while handling challenging cases of occlusion and motion blur. At the same time, the clip should be short enough to allow us to fit it at high spatial resolution in the memory of a GPU. 

The resulting clip-level instance masks ${M}^i_{t-T:t+T}$ produced densely for all overlapping clips $t=1, \hdots, L$ are then aggregated to produce video-level instance masks ${M}^i$. 

Our approach for clip-level instance segmentation is described in subsections~\ref{subsec:VideoMaskRCNN} and~\ref{subsec:MaskPropagationBranch}. We also illustrate it in Figure~\ref{high_level_arch_fig}. The subsequent clip-level instance mask aggregation method is presented in subsection~\ref{subsec:VideoMasks}.  


\begin{figure}[t]
\begin{center}
   \includegraphics[width=0.8\linewidth]{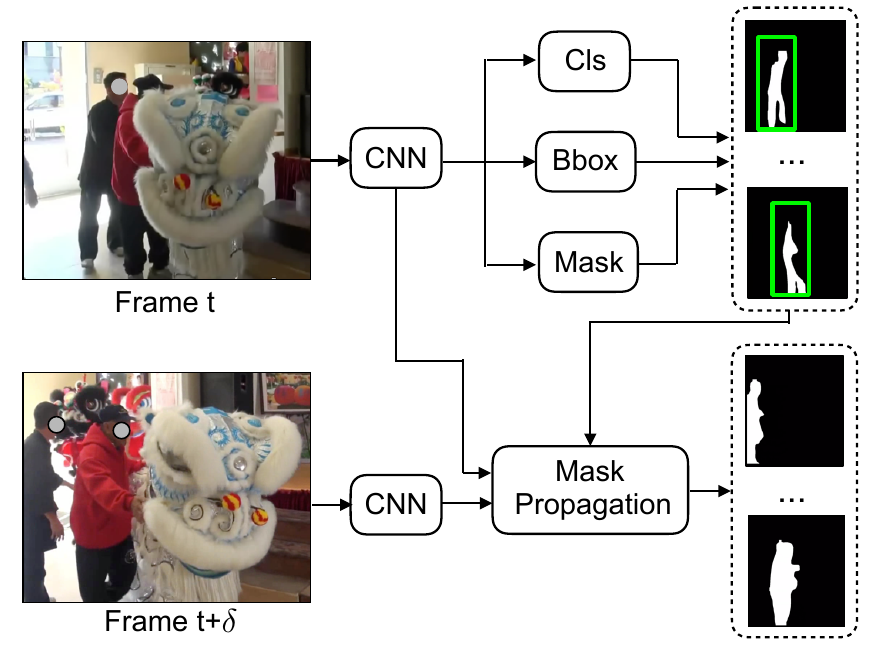}
\end{center}
\vspace{-0.4cm}
        \caption{We adapt Mask R-CNN~\cite{he2017maskrcnn} to video by adding a mask propagation branch, for tracking object segmentation instances in video. Given a video clip centered around frame $t$, our system outputs a clip-level instance segmentation track as well as a classification score and a bounding box for each object instanced detected in frame $t$. For compactness, in this figure, we illustrate our system processing a pair of frames but typically the propagation is applied from the middle frame to all the other frames in the clip.\vspace{-0.4cm}}
\label{arch_fig}
\end{figure}

\subsection{Video Mask R-CNN}
\label{subsec:VideoMaskRCNN}


Our video instance segmentation system is based on the Mask R-CNN~\cite{he2017maskrcnn} model, which we adapt to video by adding a mask propagation branch (See Figure~\ref{arch_fig}). We train our system with a multi-task loss $L_t= L_t^{cls} + L_t^{box} + L_t^{mask} + L_t^{prop}$ where $t$ denotes a time-step of a center frame. We use identical loss terms $L_t^{cls}, L_t^{box}$, $L_t^{mask}$ as in Mask R-CNN. The mask propagation loss is defined as:

\footnotesize
\begin{equation}
\label{maskprop_loss_eq}
\begin{split}
L^{prop}_t =\sum_i^{\tilde{N}_t} \sum_{t'=t-T}^{t+T} 1 - sIoU(M^i_{t-T:t+T}(t'), \tilde{M}^i_{t-T:t+T}(t'))
\end{split} 
\end{equation}
\normalsize

where $M^i_{t-T:t+T}(t') \in [0, 1]$ is the segmentation at time $t'$ for an instance $i$ predicted from a clip centered at $t$ and $\tilde{M}_{t-T:t+T}^i(t')$ is the corresponding ground truth mask at time $t'$. $\tilde{N}_t$ is the number of ground truth object instances in frame $t$, and $sIoU$ is defined as:

\begin{figure*}[t]
\begin{center}
   \includegraphics[width=0.7\linewidth]{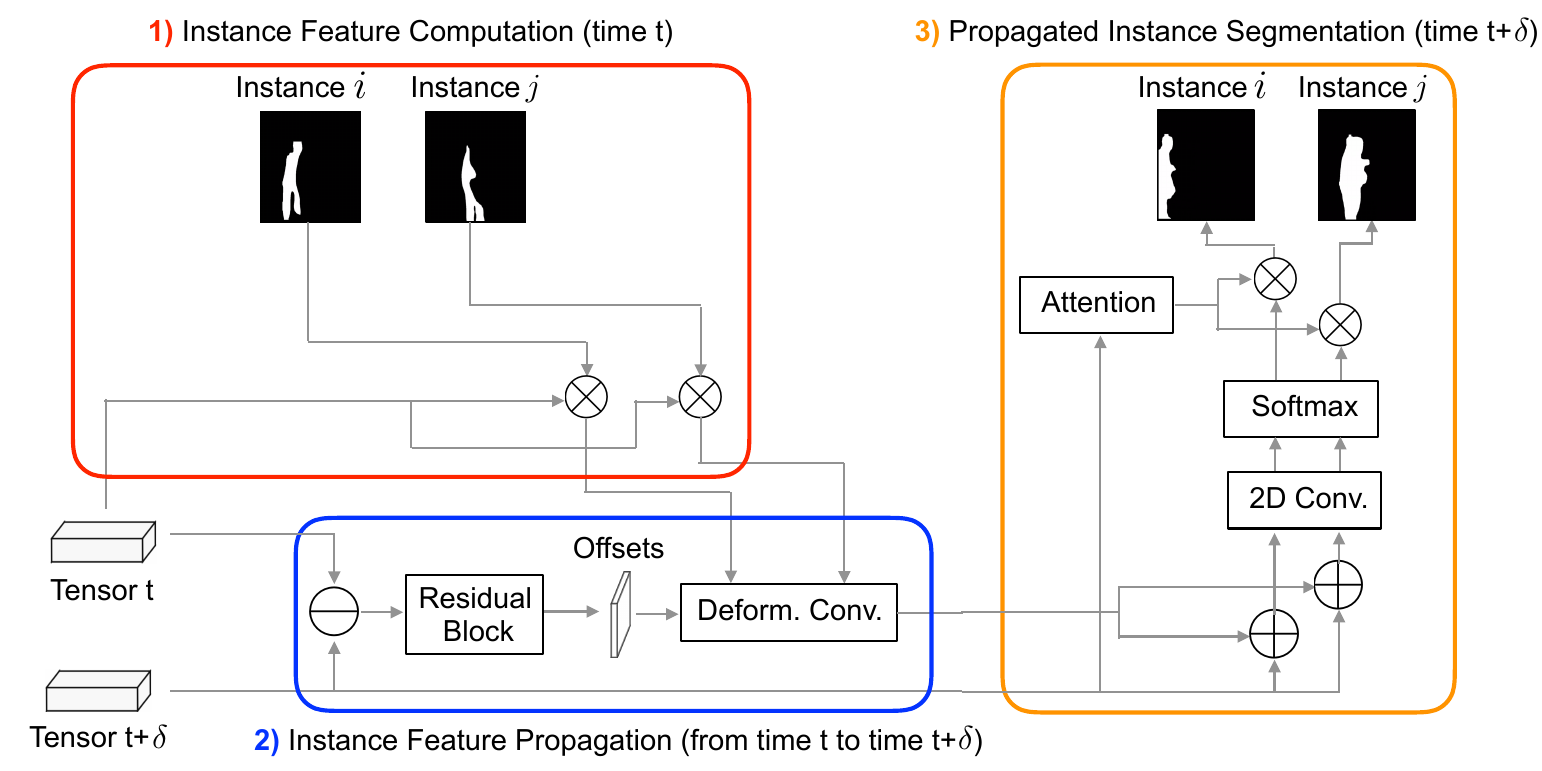}
\end{center}
\vspace{-0.4cm}
 \caption{An illustration of the 3 steps implemented by our mask propagation branch: \textbf{\color{red} 1)} For every detected instance in frame $t$, we compute an instance-specific feature tensor via element-wise multiplication between tensor $t$ and the given frame-level instance mask at frame $t$ . \textbf{\color{blue} 2)}  Next, the element-wise difference of the feature tensors associated with frames $t$ and $t+\delta$ is used to predict motion offsets between frames $t$ and $t+\delta$. The resulting offsets are used to propagate the instance-specific tensors from time $t$ to time $t+\delta$ via deformable convolution. The propagated tensors represent instance-specific features predicted for time $t+\delta$ using the tensors computed at time $t$. \textbf{\color{orange} 3)}  Lastly, we add the propagated instance feature tensors to the tensor effectively computed at $t+\delta$. A convolutional layer applied to these tensors predicts instance masks in frame $t+\delta$. The pixels that do not belong to any object instances are zeroed out using an instance-agnostic attention map. \vspace{-0.3cm}}
\label{warping_fig}
\end{figure*}



\vspace{-0.1cm}
\begin{equation}
\label{softiou_eq}
\begin{split}
sIoU(A, B) = \frac{\sum_{p}A(p)B(p)}{\sum_{p}A(p) + B(p) - A(p)B(p)}
\end{split} 
\end{equation}

where the summations in numerator and denominator are performed over every pixel location $p$. The loss above is a soft IoU loss, which we observed to work slightly better than the standard cross entropy loss for our task.




\subsection{Mask Propagation Branch}
\label{subsec:MaskPropagationBranch}


\noindent \textbf{Overview.} Our main technical contribution is the design of a mask propagation branch, that allows our method to track object instances. Given a video clip $V_{t-T:t+T}$ centered at frame $t$, our system outputs clip-level instance masks $M_{t-T:t+T}^i$ for each predicted object instance $i$ in frame $t$. Our mask propagation branch can be described in three high-level steps: 1) instance-specific feature computation, 2) temporal propagation of instance features, and 3) propagated instance segmentation. We will now describe each of these steps in more detail. We introduce our mask propagation with the example of propagating object instance masks from frame $t$ to frame $t+\delta$ where $\delta \in [-T:T]$.

\noindent \textbf{Computing Instance Specific Features.} The mask branch of our model predicts frame-level instance masks \smash{$M^i_t \in \mathcal{R}^{1 \times H' \times W'}$} from single frame inputs. We then use these frame-level instance masks to compute instance-specific features for frame $t$. Specifically, for each object instance $i$, we compute an element-wise product between \smash{$M^i_t$} and the feature tensor from the backbone network \smash{$f_t$}. This then yields a set of new feature tensors \smash{${f}^i_t \in \mathcal{R}^{C \times H' \times W'}$}, where $i=1,..,N_t$, and $N_t$ is the number of object instances detected in frame $t$. In other words, for each object instance $i$, we are zeroing out the feature values in $f_t$ that correspond to pixels not belonging to that object instance. 

\noindent \textbf{Temporally Propagating Instance Features.} Given frame-level features $f_t, f_{t+\delta} \in \mathcal{R}^{C \times H' \times W'}$ and instance-specific feature tensor \smash{${f}^i_t$}, our method generates a propagated instance feature tensor \smash{$g^i_{t,t+\delta}$}. Intuitively, \smash{$g^i_{t,t+\delta}$} represents the features predicted by our model for object instance $i$ in frame $t+\delta$ from an instance-specific feature tensor \smash{${f}^i_t$}. The tensor \smash{$g^i_{t,t+\delta}$} is generated by warping features \smash{${f}^i_t$} using the alignment computed from frame-level features $f_t$ and $f_{t+\delta}$.  We implement the propagation mechanism via a deformable convolution~\cite{8237351}, which has previously been used for aligning features computed from separate frames of a video~\cite{DBLP:conf/eccv/BertasiusTS18,NIPS2019_gberta}. Specifically, we compute the element-wise difference of tensors $f_t, f_{t+\delta}$ and feed it through a simple residual block~\cite{He2016DeepRL}, which predicts motion offsets \smash{$o_{t, t+\delta} \in \mathcal{R}^{2k^2 \times H' \times W'}$}. These offsets contain \smash{$(x,y)$} sampling locations for each entry of a $k \times k$ deformable convolution kernel~\cite{8237351}. The propagation step takes as inputs 1) the offsets \smash{$o_{t, t+\delta}$} and 2) the instance feature tensor \smash{${f}^i_t$}, and then applies deformable convolution to output the propagated instance feature tensor \smash{$g^i_{t,t+\delta}$} for each instance $i$ . We use subscript $t,t+\delta$ to denote the propagated instance feature because, although $g$ is obtained by propagating the feature tensor \smash{${f}^i_t$}, the offset computation uses both frame $t$ and frame $t+\delta$. We stress that no explicit ground truth alignment is available between frames. The deformable convolutional kernels are supervised implicitly by optimizing Eq.~\ref{maskprop_loss_eq}.


\noindent \textbf{Segmenting Propagated Instances.} Lastly, we use our propagated feature map \smash{$g^i_{t,t+\delta}$} for predicting a corresponding object instance mask in frame $t+\delta$. To do this we first, construct a new feature tensor $\phi^i_{t,t+\delta} = g^i_{t,t+\delta} + f_{t+\delta}$. 
The addition effectively overimposes the tensor $g^i_{t,t+\delta}$ predicted from time $t$ for object instance $i$ in frame $t+\delta$, with the tensor $ f_{t+\delta}$ effectively computed from the frame at time $t+\delta$. If the object instance prediction is consistent with the feature computation, the feature tensors will be aligned and thus, they will reinforce each other in the predicted region. 

Finally, the resulting feature tensors \smash{$\phi^i_{t,t+\delta}$} are fed into a $1\times1$ convolution layer that outputs instance masks for each object instance $i$ in frame $t+\delta$. The masks are normalized with the softmax nonlinearity across all $N_t$ instances. To zero-out pixels that do not belong to any object instance, we use a single $3 \times 3$ convolution that computes an instance-agnostic attention map $A_{t+\delta}$ from feature tensor $f_{t+\delta}$. We then multiply $A_{t+\delta}$ with each of our predicted instance masks. A detailed illustration of our mask propagation branch is presented in Figure~\ref{warping_fig}.


\subsection{Video-Level Segmentation Instances} 
\label{subsec:VideoMasks}




Given a video of length $L$, our goal is to produce video-level segmentation instances $M^i \in \mathcal{R}^{L \times H \times W}$. Conceptually, this requires linking clip-level instance tracks $M_{t-T:t+T}^i$ and $M_{t'-T:t'+T}^j$ when $i$ and $j$ represent the same object instance, i.e., when the instances are matching. We achieve this by assigning a video-level instance ID to each of our predicted clip-level instance tracks. Matching instance tracks are assigned the same video-level instance ID. 



\begin{figure*}[t]
\begin{center}
   \includegraphics[width=1\linewidth]{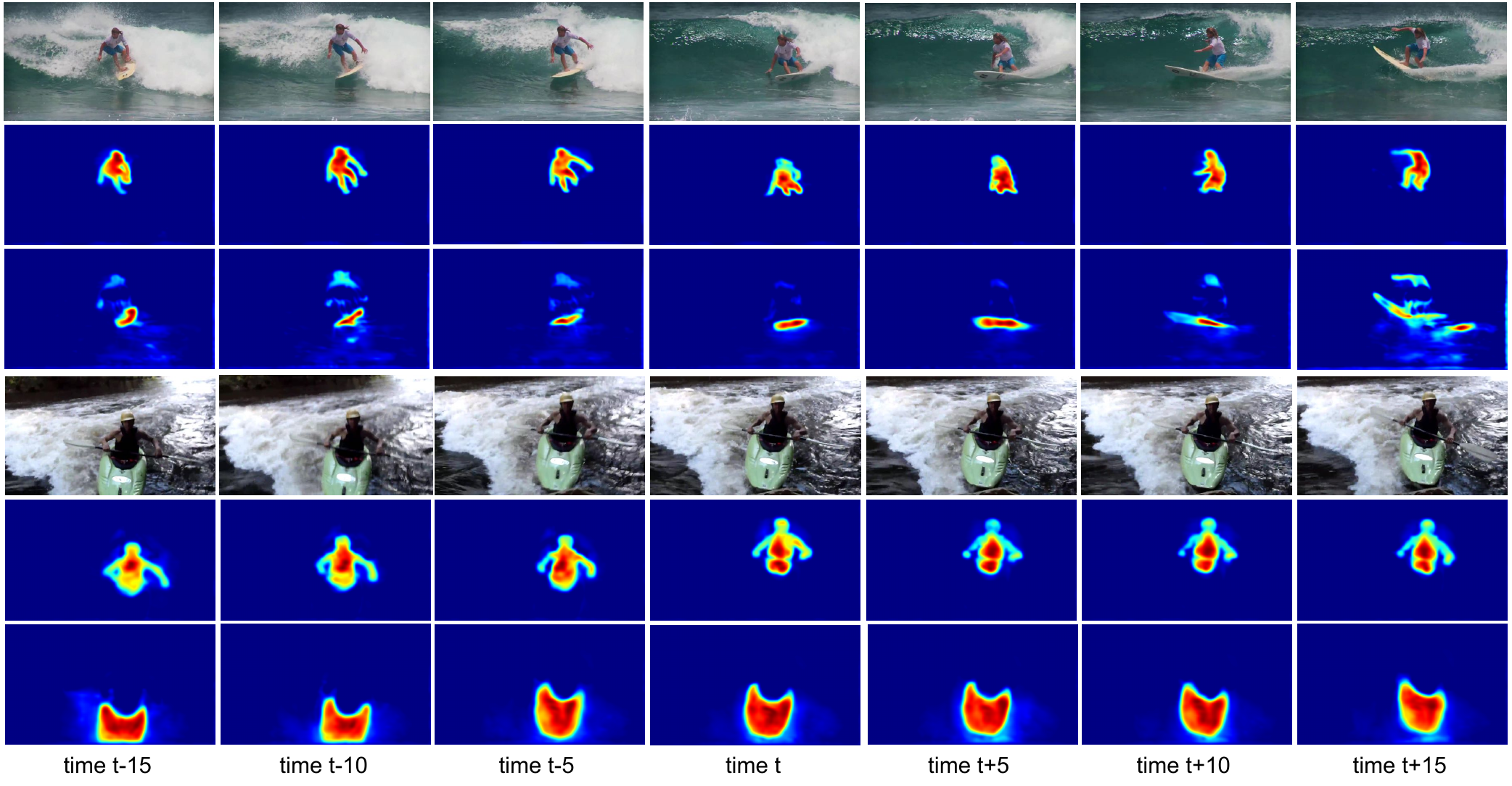}
\end{center}
\vspace{-0.5cm}
        \caption{An illustration of instance-specific features propagated from frame $t$ to other frames in the given video clip. Here, we visualize propagated activations from one randomly selected feature channel. The activations in the two rows correspond to two different object instances detected at time $t$. Our visualizations suggest that MaskProp reliably propagates features that are specific to each instance even when instances appear next to each other, and despite the changes in shape, pose and the nuisances effects of deformation and occlusion.\vspace{-0.4cm}}
\label{warped_feats_fig}
\end{figure*}

\noindent \textbf{Matching Clip-Level Instance Tracks.} Consider a pair of clip-level instance tracks \smash{$M_{t-T:t+T}^i$} and \smash{$M_{t'-T:t'+T}^j$}, that are centered around frames $t$ and $t'$ respectively. They overlap in time if $|t-t'| < 2T+1$. Let us denote their overlapping time interval as ${\cap}_{t,t'}$. Given two overlapping instance tracks, we can check if they match by comparing their predicted instance masks in the overlapping frames. If the masks of the two instance tracks align well, then they are likely to encode the same object instance. Otherwise, each track represents a different object instance. We compute a matching score $m_{t,t'}^{i,j}$ between two clip-level instance tracks using our previously defined soft IoU metric as:

\footnotesize
\begin{equation}
\begin{split}
m_{t,t'}^{i,j} =\frac{1}{|{\cap}_{t,t'}|} \sum_{\tilde{t} \in{\cap}_{t,t'}} sIoU(M^i_{t-T:t+T}(\tilde{t}), M^j_{t'-T:t'+T}(\tilde{t}))
\end{split} 
\end{equation}
\normalsize


\noindent \textbf{Video-Level Instance ID Assignment.} We denote with $\mathcal{Y}$ the set of video-level instance IDs. The set $\mathcal{Y}$ is built incrementally by matching clip-level instance tracks \smash{$M^i_{t-T:t+T}$} in order from time $t=1$ to time $t=L$. Initially, we set $\mathcal{Y} = \{1, 2, \hdots, N_1\}$ where $N_1$ is the number of object instances detected at time $t=1$ (i.e., in the first clip of the video). Let $y_t^i \in \mathcal{Y}$ denote the video-level instance ID assigned to clip-level instance track \smash{$M^i_{t-T:t+T}$}. As we move forward in time $t>1$, the video-level instance ID $y_t^i $ is assigned by matching clip-level instance \smash{$M^i_{t-T:t+T}$} to all previously processed instance tracks \smash{$M^j_{t'-T:t'+T}$} that overlap with this clip, i.e., such that $\cap_{t,t'} \neq \emptyset$. For each video-level instance ID $y \in \mathcal{Y}$ already in the ID set, we compute a score $q^i_t(y)$ capturing how well \smash{$M^i_{t-T:t+T}$} matches the tracks that have been already assigned video-level ID $y$:

\vspace{-0.2cm}
\begin{equation}
\begin{split}
q^i_t(y)=\frac{\sum_{t' \text{s.t. } \cap_{t,t'} \neq \emptyset} \sum_{j =1}^{N_{t'}}  1 \{y^j_{t'} = y\} \cdot m^{i,j}_{t,t'}}{{\sum_{t' \text{s.t. } \cap_{t,t'} \neq \emptyset}  \sum_{j =1}^{N_{t'}} 1 \{y^j_{t'} = y\}}}
\end{split} 
\end{equation}

where $1 \{y^j_{t'} = y\}$ is an indicator function that is equal to $1$ when $y^j_{t'} = y$, and $0$ otherwise. $N_{t'}$ is the number of detected instances at time $t'$, and $m^{i,j}_{t,t'}$ is the previously defined matching score between instance tracks \smash{$M_{t-T:t+T}^i$} and \smash{$M_{t'-T:t'+T}^j$}.  The score \smash{$q^i_t(y)$} is effectively an average of the soft IoU computed between instance track \smash{$M_{t-T:t+T}^i$}, and all its overlapping instance tracks that have been assigned video-level instance ID $y$.

Let \smash{$q^* = \operatorname*{max}_{y \in \mathcal{Y}}q^i_t(y)$} be the maximum score obtained by considering all possible video-level instance IDs \smash{$y \in \mathcal{Y}$}. If $q^*$ is greater than a certain threshold, the current instance track \smash{$M_{t-T:t+T}^i$} is assigned the video-level instance ID \smash{$y^* = \operatorname*{arg\,max}_{y \in \mathcal{Y}}q^i_t(y)$}. Otherwise, the clip-level track does not match any of the current video-level instances. In such case we create a new video-level instance ID and assign it to the the clip-level track while also expanding the set $\mathcal{Y}$, i.e., \smash{$y_t^i = |\mathcal{Y}| +1$} and \smash{$\mathcal{Y} = \mathcal{Y} \cup \{|\mathcal{Y}|+1\}$}.

Finally, for each video-level instance ID $y \in \mathcal{Y}$, we generate the final sequence of segmentation instance masks $M^y \in \mathcal{R}^{L \times H \times W}$ as:

\[
    M^y(t) = 
\begin{cases}
   M^i_{t-T:t+T}(t)& \text{if } y^i_t = y\\
   0              & \text{otherwise~.}
\end{cases}
\]

\begin{table*}[t]
\vspace{-0.2cm}
\setlength{\tabcolsep}{8.0pt}
\footnotesize
\begin{center}
 \begin{tabular}{c c c c c c} 
 \hline
 Method & Pre-training Data & mAP & AP@75 & AR@1 & AR@10 \\ \hline
 DeepSORT$^{\ddagger}$~\cite{DBLP:journals/corr/WojkeBP17} & Imagenet~\cite{ILSVRCarxiv14}, COCO~\cite{502} & 26.1 & 26.1 & 27.8 & 31.3\\
 FEELVOS$^{\ddagger}$~\cite{feelvos2019} & Imagenet~\cite{ILSVRCarxiv14}, COCO~\cite{502} & 26.9 & 29.7 & 29.9 & 33.4\\
 OSMN$^{\ddagger}$~\cite{DBLP:conf/cvpr/YangWXYK18} & Imagenet~\cite{ILSVRCarxiv14}, COCO~\cite{502} & 27.5 & 29.1 & 28.6 & 33.1\\
 MaskTrack R-CNN$^{\ddagger}$~\cite{Yang2019vis} & Imagenet~\cite{ILSVRCarxiv14}, COCO~\cite{502} & 30.3 & 32.6 & 31.0 & 35.5\\
 MaskTrack R-CNN$^{*}$ & Imagenet~\cite{ILSVRCarxiv14}, COCO~\cite{502} & 36.9 & 40.2 & 34.3 & 42.9\\
 EnsembleVIS~\cite{vis_challenge_winner19} & Imagenet~\cite{ILSVRCarxiv14}, COCO~\cite{502}, Instagram~\cite{mahajan2018weaklysup}, OpenImages~\cite{kuznetsova2018open} & 44.8 & 48.9 &  42.7 & 51.7\\ \hline
  MaskProp & Imagenet~\cite{ILSVRCarxiv14}, COCO~\cite{502} & 46.6 & 51.2 & 44.0 & 52.6\\
  MaskProp & Imagenet~\cite{ILSVRCarxiv14}, COCO~\cite{502}, OpenImages~\cite{kuznetsova2018open} & \bf 50.0 & \bf 55.9 & \bf 44.6 & \bf 54.5\\
  \hline
 \vspace{-0.6cm}
\end{tabular}
\end{center}
\caption{The results of video instance segmentation on the YouTube-VIS~\cite{Yang2019vis} validation dataset. We evaluate the performance of each method according to mean average precision (mAP), average precision at $75\%$ IoU threshold (AP@75), and average recall given top $1$ (AR@1) and top $10$ (AR@10) detections. The baselines denoted with $^{\ddagger}$ were implemented by the authors in~\cite{Yang2019vis}, whereas the methods marked with $^{*}$ were implemented by us, and use the same backbone and detection networks as our approach. Despite its simplicity, MaskProp outperforms all prior video instance segmentation methods by a large margin.\vspace{-0.3cm}} 
\label{results_table}
\end{table*}

\subsection{Implementation Details}
\label{sec:impl}



\noindent \textbf{Backbone Network.} As our backbone we use a Spatiotemporal Sampling Network~\cite{DBLP:conf/eccv/BertasiusTS18} based on a Deformable ResNeXt-101-64x4d~\cite{8237351,xie2016groups} with a feature pyramid network (FPN)~\cite{lin2016fpn} attached on top of it.  

\noindent \textbf{Detection Network.} For detection, we use a Hybrid Task Cascade Network~\cite{DBLP:conf/cvpr/ChenPWXLSF0SOLL19} with a $3$ stage cascade. 

\noindent \textbf{Mask Propagation Branch.} The residual block in the mask propagation branch consists of two $3 \times 3$ convolutions with $128$ output channels each. The instance feature propagation is applied to the FPN feature map with the second largest spatial resolution. To propagate instance features from one frame to another, we predict $9$ $(x,y)$ offsets for every pixel, which are then used as input to a $3 \times 3$ deformable convolutional layer with $256$ channels. To capture motion at different scales, we use three levels of dilated deformable convolutions with dilation rates $3,6,12$ as in~\cite{NIPS2019_gberta}.

\noindent \textbf{High-Resolution Mask Refinement.} Predicting masks from RoI features typically leads to low resolution predictions. We address this issue via a high-resolution mask refinement step. Given the center of a detected bounding box, we crop a $384 \times 384$ patch around the object, preserving the original aspect ratio. We then feed the RGB patch and the predicted low-resolution mask through $3$ residual blocks each with $128$ channels to obtain a high-resolution mask. 

\noindent \textbf{Scoring Video-Level Sequences.} Each video-level sequence contains a list of classification scores and predicted object labels. To assign a confidence score to each video-level sequence, we average classification scores associated with that sequence (separately for each object category).  


\noindent \textbf{Training.} We use a similar training setup as in~\cite{DBLP:conf/cvpr/ChenPWXLSF0SOLL19}.  The loss weights for each of the three cascade stages are set to $1, 0.5,$ and $0.25$ respectively. The loss weight for the semantic segmentation branch is set to $0.1$. We train our model on pairs of frames, where the second frame in a pair is randomly selected with a time gap $\delta \in  [-25, 25]$ relative to first frame. We use a multi-scale training approach implemented by resizing the shorter side of the frame randomly between $400$ and $800$ pixels. Our model is trained in a distributed setting using $64$ GPUs, each GPU holding a single clip. The training is done for $20$ epochs with an initial learning rate of $0.008$, which is decreased by $10$ at $16$ and $19$ epochs. We initialize our model with a Mask R-CNN pretrained on COCO for the instance segmentation. The hyperparameters of RPN and FPN are the same as in~\cite{DBLP:conf/cvpr/ChenPWXLSF0SOLL19}.


\noindent \textbf{Inference.} During testing, we run the bounding box prediction branch on $1000$ proposals, apply non-maximum suppression, and use boxes with a score higher than $0.1$ as input to the mask prediction and mask propagation branches. During inference, our MaskProp is applied to video clips consisting of $13$ frames. 



\section{Experimental Results}

In this section, we evaluate MaskProp for video instance segmentation on YouTube-VIS~\cite{Yang2019vis}, which contains $2,238$ training, $302$ validation, and $343$ test videos. Each video is annotated with per-pixel segmentation, category, and instance labels. The dataset contains $40$ object categories. Since the evaluation on the test set is currently closed, we perform our evaluations on the validation set.

\begin{figure}[t]
\begin{center}
   \includegraphics[width=0.69\linewidth]{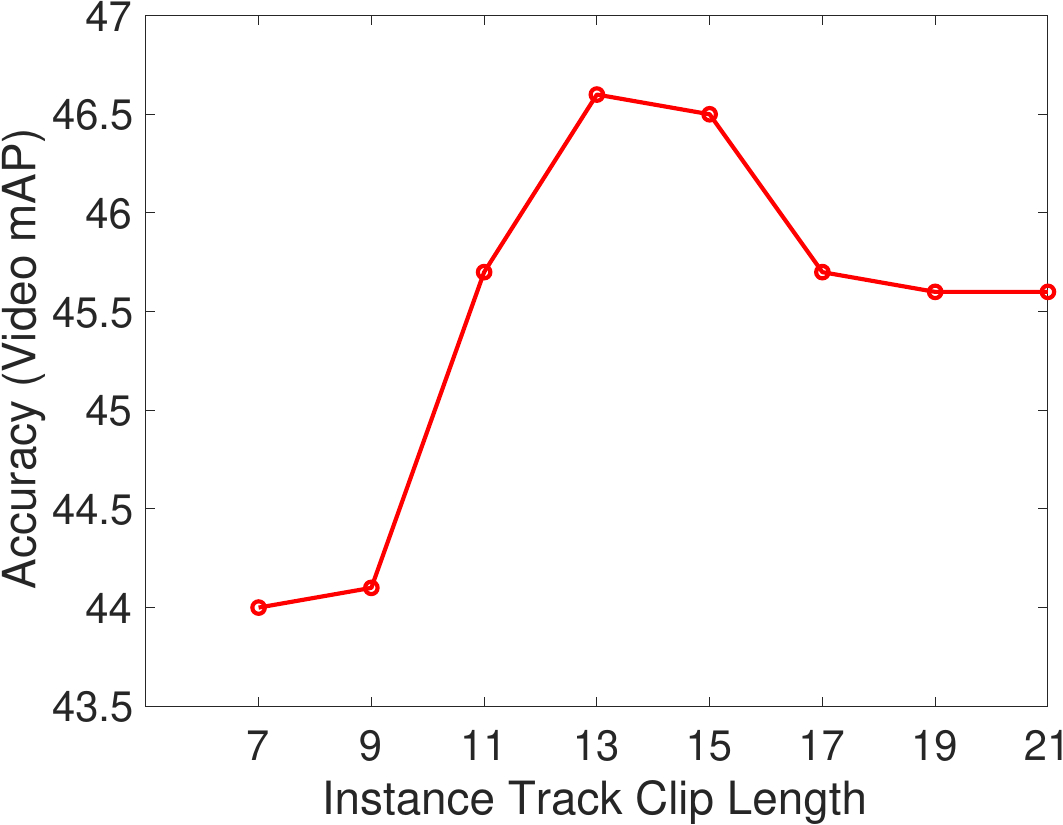}
\end{center}
\vspace{-0.4cm}
        \caption{ We plot video mAP as a function of an instance track clip length (denoted as $2T+1$ in the paper). Based on these results, we observe that optimal video instance segmentation performance is achieved when we propagate instance masks to $T=6$ previous and subsequent frames.\vspace{-0.4cm}}
\label{ablation_fig}
\end{figure}





\subsection{Quantitative Results}

Video instance segmentation is a very recent task~\cite{Yang2019vis}, and thus, there are only a few established baselines that we can compare our work to. We include in our comparison MaskTrack R-CNN~\cite{Yang2019vis}, and the EnsembleVIS method, which won the ICCV 2019 video instance segmentation challenge~\cite{vis_challenge_winner19}.  Additionally, to make the comparison with MaskTrack R-CNN more fair, we reimplement it using the same backbone and detection networks as our MaskProp (see MaskTrack R-CNN* in Table~\ref{results_table}).  

We present our quantitative results in Table~\ref{results_table}, where we assess each method according to 1) mean video average precision (mAP), 2) video average precision at IoU threshold of $75\%$, and 3) average recall given $1$ and $10$ highest scored instances per video. From these results, we observe that our MaskProp outperforms all the other baselines according to all four evaluation metrics, thus achieving state-of-the-art results in video instance segmentation on YouTube-VIS. It can be noted that we outperform EnsembleVIS~\cite{vis_challenge_winner19} by $1.8$ mAP even though our approach is much simpler and even if it uses orders of magnitude less labeled data for pre-training. Furthermore, compared to the MaskTrack R-CNN, our method achieves a $16.3$ improvement in mAP. We also note that our implementation of MaskTrack~\cite{Yang2019vis} significantly improves upon the original work, but it is still $9.7$ worse in mAP compared to our model. Lastly, we note that additionally pre-training MaskProp on the OpenImages~\cite{kuznetsova2018open} dataset as was done in~\cite{vis_challenge_winner19} further boosts our performance to $50.0$ mAP.

\begin{table}[t]
\setlength{\tabcolsep}{10.0pt}
\small
\begin{center}
 \begin{tabular}{c c c} 
 \hline
Method & mAP & AP@75\\ \hline
FlowNet2 Propagation & 31.4 & 33.6\\
MaskTrack R-CNN$^{*}$ & 36.9 & 40.2\\
MaskProp & \bf 46.6 &  \bf 51.2\\
  \hline
 \vspace{-0.7cm}
\end{tabular}
\end{center}
\caption{Here, we study the effectiveness of our mask propagation branch.  If we replace it with the FlowNet2 propagation scheme, where masks are propagated using the optical flow predicted by a FlowNet2 network~\cite{DBLP:journals/corr/IlgMSKDB16}, the accuracy drops from $46.6$ mAP to $31.4$ mAP. Similarly, if we replace our mask propagation branch with the tracking branch from MaskTrack R-CNN, the accuracy drops to  $36.9$ mAP. Note that all of these baselines are implemented using the same backbone and detection networks. \vspace{-0.3cm}}
\label{maskprop_abl_table}
\end{table}

%
%
%
%




\subsection{Ablation Experiments}



\noindent  \textbf{Mask Propagation Branch.} To investigate the effectiveness of our mask propagation branch, we compare our method with a FlowNet2~\cite{DBLP:journals/corr/IlgMSKDB16} propagation baseline. For this baseline, we use exactly the same setup as for our MaskProp, except that instance masks are propagated using the optical flow predicted by a FlowNet2 network~\cite{DBLP:journals/corr/IlgMSKDB16} rather than our proposed mask propagation scheme. For a more complete comparison, we also include the MaskTrack R-CNN$^{*}$ from Table~\ref{results_table}, which uses the originally proposed tracking branch~\cite{Yang2019vis}, but is implemented using the same backbone and detection networks as our MaskProp. 

These baselines allow us to directly compare the effectiveness of our mask propagation scheme versus the propagation mechanisms employed by FlowNet2 and MaskTrack R-CNN~\cite{Yang2019vis} methods. The results in Table~\ref{maskprop_abl_table} show that MaskProp outperforms these baselines by a large margin. 

\noindent  \textbf{Instance Track Clip Length.} Due to occlusions, object instances in video may not be visible in some frames. If there are occlusions separated by $2T'+2$ time-steps, we can use $T > T'$ to predict longer clip-level instance tracks. In Figure~\ref{ablation_fig}, we study video instance segmentation performance as a function of instance track clip length (denoted as $2T+1$ in the paper). Our results indicate that the best accuracy is achieved when we use a clip length of $13$, meaning that we propagate instances to $6$ previous and $6$ subsequent frames. 


\noindent  \textbf{High-Resolution Mask Refinement.} We also study the impact of our high-resolution mask refinement, described in Subsection~\ref{sec:impl}. We report that removing this refinement causes a drop of $1.9\%$ in video instance segmentation mAP.

\begin{table}[t]
\setlength{\tabcolsep}{10.0pt}
\small
\begin{center}
 \begin{tabular}{c c c} 
 \hline
Backbone Network & mAP & AP@75\\ \hline
ResNet-50 & 40.0 & 42.9\\
ResNet-101 & 42.5 & 45.6\\
ResNeXt-101-64x4d & 44.3 & 48.3\\
STSN~\cite{DBLP:conf/eccv/BertasiusTS18}-ResNeXt-101-64x4d & \bf 46.6 &  \bf 51.2\\
  \hline
 \vspace{-0.7cm}
\end{tabular}
\end{center}
\caption{We study the effect of frame-level instance masks to our system's performance. We evaluate our method's accuracy when using instance masks obtained from Mask R-CNN with several different backbones. Our results indicate that frame-level instance masks obtained from stronger models lead to better video instance segmentation results. \vspace{-0.4cm}}
\label{backbone_table}
\end{table}

\noindent  \textbf{Importance of Frame-Level Instance Masks.} As described in our main draft, we use frame-level instance masks for instance-specific feature computation. To investigate the contribution of these masks to the performance of our system, we experiment with masks obtained from several different Mask R-CNN models. In Table~\ref{backbone_table}, we present our results for this ablation. Our results indicate that frame-level instance masks obtained from stronger models allow us to achieve better video instance segmentation performance. Thus, we expect that future improvements in image instance segmentation will further benefit our method.


\subsection{Qualitative Results}

\begin{figure*}[t]
\begin{center}
   \includegraphics[width=0.85\linewidth]{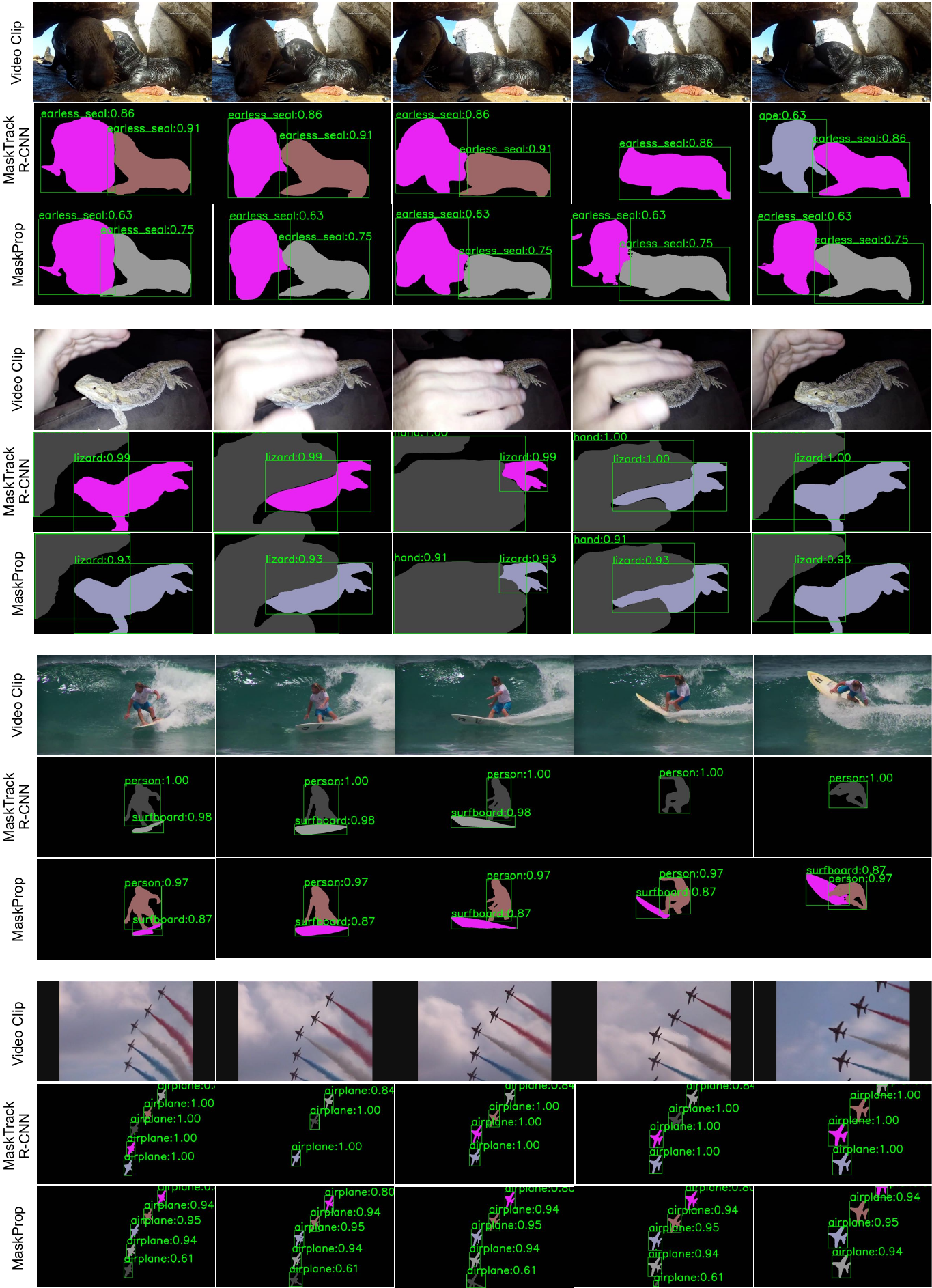}
\end{center}
\vspace{-0.4cm}
       \caption[Caption for LOF]{We compare our video instance segmentation results with MaskTrack R-CNN~\cite{Yang2019vis} predictions. Different object instances are encoded with different colors. The first row for each video shows the original frames. The second row illustrates the mask predictions of MaskTrack R-CNN and the third row those obtained with our MaskProp. Compared to MaskTrack R-CNN, our MaskProp tracks object instances more robustly even when they are occluded or overlap with each other. \vspace{-0.4cm}}
\label{results_fig}
\end{figure*}

In Figure~\ref{results_fig}, we compare our predicted clip-level instance tracks (last row of predictions for each clip) with the MaskTrack R-CNN predictions (first row of predictions). We use different colors to represent different object instances. Our qualitative results suggest that our MaskProp produces more robust and temporally coherent instance tracks than MaskTrack R-CNN. Such differences in performance are especially noticeable when a video contains large object motion, occlusions, or overlapping objects. 





In Figure~\ref{warped_feats_fig}, we also visualize instance-specific features that are propagated from frame $t$ to other frames in the given video clip for two different object instances detected in frame $t$. Here, we show activations from a randomly selected feature channel. Based on these results, we observe that our MaskProp reliably propagates features that are specific to each instance despite motion blur, object deformations and large variations in object appearance.




%

\section{Conclusion}


In this work, we introduced MaskProp, a novel architecture for video instance segmentation. Our method is conceptually simple, it does not require large amounts of labeled data for pre-training, and it produces state-of-the-art results on YouTube-VIS dataset. In future, we plan to extend MaskProp to scenarios where only bounding box annotations are available. We are also interested in applying our method to problems such as pose estimation and tracking.

{\small
\bibliographystyle{ieee}
\bibliography{gb_bibliography}
}

\end{document}